\documentclass[final,5p,times]{elsarticle}

\usepackage{tikz}
\usepackage{algorithm}
\usepackage{algorithmic}
\usepackage{hyperref}
\usepackage[utf8]{inputenc}
\usepackage[T1]{fontenc}
\usepackage{color,soul}
\usepackage{times}
\usepackage{amssymb} 
\usepackage[nointegrals]{wasysym} 
\usepackage{xcolor}
\usepackage{epsfig}
\usepackage{graphicx}
\usepackage{multicol,multirow}
\usepackage{booktabs}
\usepackage{bigstrut}
\usepackage{rotating}
\usepackage{adjustbox}
\usepackage{mathtools}
\usepackage{subfigure}
\usepackage{longtable}
\usepackage{array}
\usepackage{amsmath,amssymb,amsfonts}
\usepackage{algorithmic}
\usepackage{graphicx}

\usepackage{graphicx,subfigure}
\graphicspath{ {./images/} }
\usepackage{booktabs}
\usepackage{adjustbox}
\usepackage{tablefootnote}
\usepackage{multirow}
\usepackage{textcomp}
\usepackage{xcolor}

\journal{Arxiv}

\begin{document}

\begin{frontmatter}



\title{LSSF-Net: Lightweight Segmentation with Self-Awareness, Spatial Attention, and Focal Modulation}

\author[inst1]{Hamza Farooq}
\author[inst1]{Zuhair Zafar}
\author[inst1]{Ahsan Saadat}
\author[inst3]{Tariq M Khan}
\author[inst2]{Shahzaib Iqbal}
\author[inst3]{Imran Razzak}

\affiliation[inst1]{School of Electrical Engineering and Computer Science (SEECS), National University of Sciences & Technology (NUST), Islamabad 44000, Pakistan}

\affiliation[inst2]{Department of Electrical Engineering, Abasyn University Islamabad Campus, Pakistan}
\affiliation[inst3]{School of Computer Science and Engineering, UNSW, Sydney, Australia}

\begin{abstract} Accurate segmentation of skin lesions within dermoscopic images plays a crucial role in the timely identification of skin cancer for computer-aided diagnosis on mobile platforms. However, varying shapes of the lesions, lack of defined edges, and the presence of obstructions such as hair strands and marker colors make this challenge more complex. \textcolor{red}Additionally, skin lesions often exhibit subtle variations in texture and color that are difficult to differentiate from surrounding healthy skin, necessitating models that can capture both fine-grained details and broader contextual information. Currently, melanoma segmentation models are commonly based on fully connected networks and U-Nets. However, these models often struggle with capturing the complex and varied characteristics of skin lesions, such as the presence of indistinct boundaries and diverse lesion appearances, which can lead to suboptimal segmentation performance.To address these challenges, we propose a novel lightweight network specifically designed for skin lesion segmentation utilizing mobile devices, featuring a minimal number of learnable parameters (only 0.8 million). This network comprises an encoder-decoder architecture that incorporates conformer-based focal modulation attention, self-aware local and global spatial attention, and split channel-shuffle. The efficacy of our model has been evaluated on four well-established benchmark datasets for skin lesion segmentation: ISIC 2016, ISIC 2017, ISIC 2018, and PH2. Empirical findings substantiate its state-of-the-art performance, notably reflected in a high Jaccard index. \end{abstract}


\end{frontmatter}

\section{Introduction}
In an era marked by an ever-growing concern for public health, the spectre of skin cancer emerges as a subject of paramount importance, demanding our attention and understanding. Medical images, which play an important role in the process of diagnosis and treatment by physicians \cite{khan2019boosting,khan2020exploiting,khan2021residual,iqbal2022g, khan2023retinal}, have become particularly vital for current vision tasks on medical images, highlighting the critical role of accurate skin lesion segmentation. Among the myriad forms of skin cancer, melanoma emerges as the most formidable adversary, with the potential to be life-threatening. The linchpin in the battle against this risk is early detection, which is proven to be a critical factor in ensuring effective treatment and ultimately the survival of patients. It is abundantly clear that the sooner skin lesions are pinpointed, the greater the opportunity for patients to receive precisely tailored care, markedly improving their prospects of a successful recovery. Melanoma, in particular, presents itself through pigmented lesions that grace the surface of the skin, making it a prime candidate for early identification, thanks to the intelligent discernment of healthcare professionals. However, the labyrinth of skin cancer diagnosis remains a formidable challenge for dermatologists, primarily due to the immense diversity of skin lesions and the complicated task of distinguishing between benign and malignant growths.

In recent times, deep learning, especially harnessing the powerful features extraction capabilities of convolutional neural networks (CNN) \cite{iqbal2022recent,iqbal2023ldmres,javed2024advancing,khan2024lmbf,iqbal2024euis}, has made significant strides in the domain of medical image segmentation \cite{iqbal2022g, iqbal2023ldmres, abdullah2021review, imtiaz2021screening,khan2021residual,khan2022leveraging,arsalan2022prompt,khan2022mkis,khan2023simple,naqvi2023glan}. This development has led to substantial improvements in the precision of medical image segmentation tasks. The CNN framework, which consists of convolutional and down-sampling layers, operates on the principle that lower convolutional layers offer a more localised perspective and finer location information, while higher convolutional layers provide broader contextual insight into the entire image \cite{ioffe2015}, essential for segmentation tasks. 
In light of these advances, numerous models based on the full convolutional network (FCN) have been introduced to improve image segmentation \cite{asadi2020multi}. In particular, the structure of the encoding and decoding network, as epitomised by U-Net \cite{ronneberger2015u,qayyum2023two}, mitigates the loss of fine-grained details caused by multiple downsampling steps by incorporating skip connections between the encoder and the decoder, thus amplifying the performance of the network. This underscores the effectiveness of the encode-decode network architecture. Subsequently, various networks following U-shaped structures, including Res-UNet \cite{xiao2018weighted} and Attention R2U-Net \cite{alom2018recurrent}, were proposed. However, these models still faced the challenge of effectively extracting and using multiscale contextual features within a single stage. This limitation was particularly relevant in the realm of medical images, where the target regions often closely resembled their surroundings, necessitating the consideration of broader contextual information to avoid ambiguous decisions. 

To address this, researchers have devised methodologies to incorporate multiscale information, such as PSPNet \cite{zhao2017pyramid}, PoolNet \cite{liu2019simple}, DeepLabV3 \cite{chen2017rethinking}, and CE-Net \cite{gu2019net}. These approaches focus primarily on processing high-level feature information while downplaying location-based detail information present in low-level feature information. Although CNN-based methods excel in feature extraction, they tend to fall short of capturing long-distance dependencies due to the inherent limitations of convolution operations \cite{schlemper2019attention}. Consequently, these methods often struggle with target areas that exhibit substantial variations in texture, size, and shape.

 In response, some researchers have introduced attention mechanisms into CNNs to overcome this limitation \cite{xing2022cm}. Furthermore, the successful integration of Transformers into computer vision has opened new avenues \cite{dosovitskiy2020image}. Transformers operate on a sequence-to-sequence prediction architecture, circumventing the need for convolution operators and relying solely on self-attentive mechanisms to extract information about image characteristics, allowing the establishment of effective long-range dependencies.
 
Transformers have consistently demonstrated their ability to match or surpass state-of-the-art performance in various vision tasks. These models excel in capturing global context, but their effectiveness in capturing fine-grained details, especially in the case of medical images, is limited. They lack built-in spatial bias when it comes to modelling local information. Furthermore, transformer-based network structures are highly dependent on large datasets for optimal performance \cite{zheng2021rethinking}. Here, the CNN architecture proves to be a valuable counterpart, effectively compensating for these limitations. 

Recent research has explored the fusion of CNNs with Transformers for medical image segmentation. Models such as TransUNet \cite{chen2021transunet} and subsequent studies \cite{zhang2021transfuse,chen2023transattunet} have used CNNs as the foundational network, and Transformers facilitate long-range dependencies on high-level features. However, these approaches often overlook the valuable spatial information present in shallow networks, concentrating on context modelling at a single scale, disregarding cross-scale dependencies and consistency. Some scholars argue that employing just one or two layers of Transformers \cite{zhou2021nnformer} fails to combine convolutional representations that depend on CNNs for long-distance relationships. 

This paper introduces an innovative lightweight network structure, termed LSSF-Net, specifically designed for the segmentation of skin lesions and the analysis of medical images within computer-aided diagnosis (CAD) systems. The proposed model builds on the well-established encoding-decoding network architecture, specifically using the lightweight T-Net-based model \cite{khan2022t}, which is known for its efficiency and effectiveness in medical image segmentation. Building on this foundation, our LSSF-Net incorporates several key enhancements to significantly improve feature extraction. These enhancements include a novel booster architecture, self-aware local and global spatial attention (SAB), normalised focal modulation-based skip connections (CFMA) and a split channel shuffle mechanism (SCS). Together, these innovations improve the model's ability to capture fine-grained details and global context, effectively addressing the challenges posed by the complex nature of medical images. The LSSF-Net is designed to deliver high accuracy and efficiency while maintaining a lightweight structure, making it highly suitable for deployment on mobile devices with limited computational power. This work represents a significant advancement in the field by offering a solution that balances top-tier performance with resource efficiency, providing an effective and accessible tool for medical image analysis in resource-constrained environments.

The backbone of the introduced LSSF-Net consists of two parallel branches of Convolutional Neural Networks (CNNs) and a booster architecture.  CNNs focus on extracting multiscale feature information from the original input image, while the Booster concurrently models global contextual information to establish long-range dependencies. Recognising the computational cost associated with high-level semantic features, the model strategically maximises the retention of location information within low-level semantic features, as they contribute less to network performance. This thoughtful consideration aims to optimise computational efficiency without compromising overall segmentation quality \cite{khan2022t}.

For the decoding component, the same encoder structure is employed, and a Conformer-based Focal Modulation Attention (CFMA) is introduced as a skip connection from the encoder booster to the decoder. This addition enhances the acquisition of detailed global and local feature information during the decoding phase. Furthermore, to intensify interconnections between decoder blocks, facilitating dense links that improve feature preservation during the upsampling process, transformer-based attention (TA) is employed at the bottleneck of feature enhancement.

The main contributions of this work can be summarised as follows.

\begin{enumerate}
    \item \textbf{Novel Architecture:}  The proposed medical segmentation model introduces a novel architecture that features a parallel booster encoder and decoder model. This design facilitates the extraction of all feature sets and improves the segmentation capabilities.
    
    \item \textbf{Enhanced Feature Information:} To obtain more detailed global and local feature information, focal modulation is coupled with conformer attention at the skip connection. This modification aims to improve the model's ability to capture intricate details and contextual information.
    
    \item \textbf{Dense Interconnections:} The model intensifies the interconnections between decoder blocks, establishing dense links to facilitate the preservation of improved features during the crucial up-sampling process. This contributes to maintaining the integrity of features across different scales.
    
    \item \textbf{Transformer-Based Attention:} To improve features at the bottleneck, transformer-based attention is strategically used. This, combined with special enhancements to local-global characteristics, ensures that essential information is retained and utilised effectively during the segmentation process.
    
    \item \textbf{Validation and Comparison:} The proposed network's robustness and generality are validated through comprehensive comparisons with the current popular methods. This comparative analysis aims to showcase the efficacy and competitive performance of the model in the domain of medical image segmentation.
\end{enumerate}

\begin{figure*}[h]
    \centering
    \includegraphics[width=\textwidth]{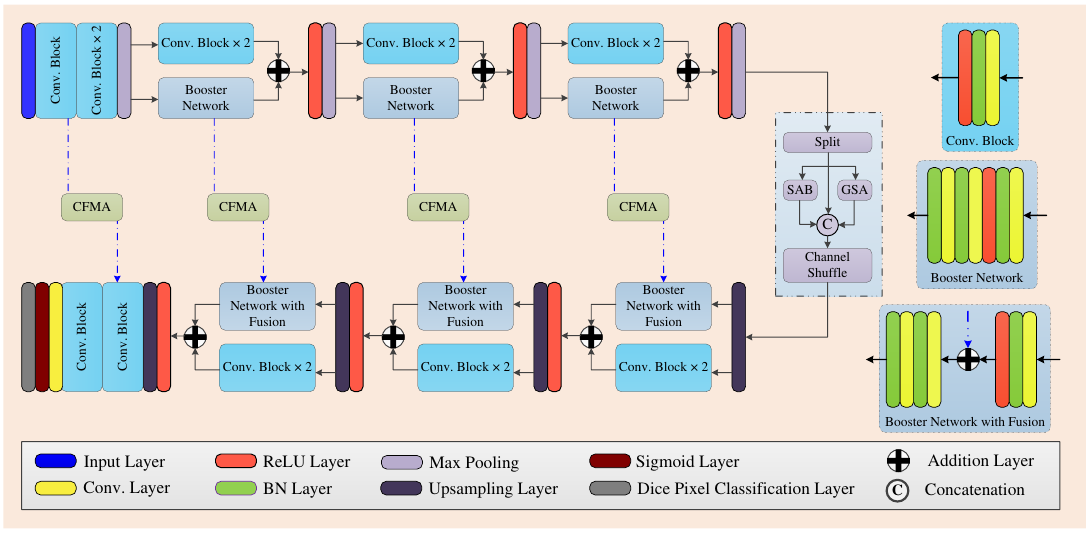}
    \caption{Block diagram of the proposed LSSF-Net. ``CFMA'' is conformer-based focal modulation attention, ``SAB'' is the self-attention block, and ``GSA'' is global spatial attention.}
    \label{fig:model}
\end{figure*} 

\section{Literature Review}

In the modern world, deep learning-based methods demonstrate better performance in the realm of medical segmentation, particularly in tasks such as segmentation of skin lesions \cite{sarker2018slsdeep}. These methods automatically extract features from the dataset and exhibit greater robustness compared to conventional hand-crafted feature extraction techniques. Ever since the introduction of UNet \cite{ronneberger2015u}, its encoder-decoder architecture has emerged as the dominant method in medical segmentation. UNet efficiently incorporates basic feature information by establishing a direct connection between the encoder and the decoder. According to a survey \cite{mirikharaji2022deep}, 87.2


\subsection{UNet based Segmentation}
In the modern era of medical image analysis, deep learning-based methods have showcased remarkable performance, particularly in tasks such as segmentation of skin lesions \cite{sarker2018slsdeep}. Among these methods, UNet and its variants have emerged as dominant players \cite{ronneberger2015u} shown in the figure. UNet adopts an encoder-decoder architecture with skip connections, enabling efficient feature extraction and preservation of detailed information. Over time, several enhancements have been proposed to the original UNet architecture, each with the aim of improving segmentation accuracy and robustness.
For example, Res-UNet \cite{xiao2018weighted} integrates residual structures in both the encoding and decoding stages, improving the retention of detailed information. UNet++ \cite{zhou2018unet++} takes a different approach by incorporating dense connections of residual structures, facilitating the accumulation of multiscale feature information. Attention mechanisms, widely successful in natural image processing, have found increasing application in medical segmentation tasks, yielding satisfactory results. Notable approaches include Attention R2U-Net \cite{alom2018recurrent}, which combines residual and recurrent networks with attention gates to improve focus, and MCGUNet \cite{asadi2020multi}, incorporating SE modules and bidirectional ConvLSTM in skip connections for dynamic feature adjustment.

\subsection{Attention Mechanisms in Medical Image Segmentation}
Researchers have proposed innovative techniques to refine skip connection feature maps, leveraging attention mechanisms to improve segmentation performance. One such approach involves the inclusion of a spatial enhancement module within skip connections, which facilitates the representation of crucial spatial details for semantic segmentation. By integrating this module, the network effectively captures and leverages spatial information, leading to better segmentation performance.
The Attention U-Net architecture \cite{oktay2018attention}  represents a significant advancement in this domain, incorporating attention gates within skip connections to address semantic ambiguity between encoder and decoder layers. Using attention gates, the model can selectively emphasise certain features of the encoder, providing better guidance and focus during the decoding process. This enables the model to capture relevant information more effectively, ultimately improving the results of the segmentation.

\subsection{Transformer Based Segmentation}
The transformative impact of Vision Transformers (ViT), as introduced by \cite{dosovitskiy2020image}, marked a significant milestone in the field of computer vision by bringing transformers, originally designed for sequential data processing, into the realm of visual tasks. ViT demonstrated remarkable performance, leveraging the transformer's capacity to capture global dependencies within images.
Building upon ViT's success, subsequent advancements in vision tasks have blossomed, inspired by its pioneering approach. For instance, DeiT \cite{touvron2022deit} explored efficient training strategies tailored to ViT architectures, enhancing scalability and performance. PVT (Pyramid Vision Transformer) \cite{dong2021polyp} introduced a pyramid transformer with Shifted Relative Attention (SRA) mechanisms, reducing computational complexity while preserving effectiveness.
The Swin Transformer \cite{liu2021swin}, represents another notable stride in hierarchical vision transformers. Its innovative window-based mechanism enhances feature locality, addressing limitations observed in previous transformer architectures. Moreover, transformers have found applications in various specific tasks within computer vision. SETR (Semantic Segmentation Transformer) leverages transformers for semantic segmentation, with ViT serving as a backbone architecture. SegFormer, introduced by Xie et al. \cite{xie2021segformer}, offers a straightforward and efficient design for semantic segmentation, powered by transformer architectures. Furthermore, Uformer, as proposed by Wang et al. \cite{wang2022uformer}, introduces a general U-shaped transformer architecture tailored for image restoration tasks, showcasing the versatility of transformer-based approaches across a wide range of applications within computer vision.
These developments underscore the transformative potential of transformers in reshaping the landscape of computer vision tasks, offering novel solutions and insights into addressing complex visual challenges. As researchers continue to innovate and refine transformer-based architectures, the future holds promising prospects for further advancements in visual understanding and processing.
\subsection{Hybrid Transformers and UNet-based Segmentation}
With the rise of Transformers as a powerful tool in computer vision, their integration into medical segmentation has attracted significant attention from researchers, showing promising results. In particular, TransUNet \cite{chen2021transunet}, is a trailblazer in incorporating Transformers into medical segmentation tasks. This pioneering methodology merges the UNet encoder with Transformer architecture, diverging from traditional image-based input methods by operating on high-level features.
The innovative fusion of UNet and Transformers in TransUNet marks a departure from conventional approaches, offering a fresh perspective on medical image segmentation. By leveraging the strengths of both architectures, TransUNet capitalizes on the hierarchical representations learned by the UNet encoder and the attention mechanisms of Transformers. This synergy enables the model to capture intricate spatial dependencies within medical images effectively, leading to improved segmentation performance.
As researchers continue to explore the potential of Transformers in medical imaging tasks, TransUNet serves as a foundational framework, inspiring further advancements and innovations in the intersection of Transformer-based methods and medical segmentation techniques.
TransFuse \cite{zhang2021transfuse} offers a fresh perspective by bridging CNNs and Transformers in parallel, presenting a novel approach in the domain of medical segmentation. Central to its innovation is the introduction of the BiFusion fusion module, which adeptly combines shallow network features from CNN encoders with feature information extracted via Transformers. This integration facilitates a comprehensive understanding of the input data, leveraging the strengths of both architectures to enhance segmentation accuracy.\\
In contrast, TransAttunet \cite{chen2023transattunet} introduces the Self-Aware Attention (SAA) module, a novel mechanism that merges Transformer Self-Attention (TSA) and Global Spatial Attention (GSA), fundamental components of Transformer architecture. By incorporating these attention mechanisms, TransAttunet efficiently captures non-local interactions among encoder features, thereby enriching the segmentation process. However, despite these advances, there remains a challenge in fully harnessing the richness of feature information across multiple scales.
The quest for establishing long-range dependencies using Transformers has been transformative in medical segmentation tasks. However, exploring feature information on different scales remains an ongoing pursuit, highlighting the need for further research and innovation to leverage the full potential of hybrid CNN-Transformer architectures.

The integration of attention mechanisms and transformers has significantly advanced skin lesion segmentation in prior research. However, it is crucial to acknowledge the limitations of these earlier methods. Although they incorporate attention mechanisms, they often struggle to effectively merge spatial and channel information, which can impact precision. Furthermore, transformer-based models in this context have primarily focused on long-range dependencies, potentially missing the finer details essential for accurate skin lesion segmentation. Lightweight models, when combined with suitable attention mechanisms and feature enhancements, demonstrate superior performance by striking a balance between model complexity and precision. This amalgamation ensures that the crucial finer details necessary for accurate segmentation are preserved, offering a promising avenue to advance skin lesion analysis and achieve superior segmentation results.

\section{Proposed Methodology}
In this section, we will briefly discuss the architecture of the proposed LSSF-Net. Fig/ \ref{fig:model} presents the block diagram of the proposed model, which consists of four encoder-decoder blocks, conformer-based focal modulation attention (CFMA) blocks in skip connections, self-attention block (SAB) and global spatial attention (GSA) blocks in the bottleneck layer of the proposed LSSF-Net. Details for each component are provided in the following subsections.

\subsection{Model Architecture}

In the proposed implementation, we have employed four encoder-decoder blocks. Let $l^{n\times n}$ be the $n\times n$ convolution operation $f^{n\times n}$ followed by batch normalisation ($\beta _{n}$) and ReLU ($\Re$) operations for any given input ($\texttt{In}$) as defined by (Eq. \ref{Eq1}).
\begin{equation}
l^{n\times n} = \Re \left ( f^{n\times n}\left ( \texttt{In} \right ) \right )
    \label{Eq1}
\end{equation}

The initial skip connection ($s_{o}$) is computed by applying the $l^{3\times 3}$ operation to the input of the network ($X_{in}$) as shown in (Eq. \ref{Eq2}).

\begin{equation}
    s_{o}=l^{3\times 3}(X_{in})
    \label{Eq2}
\end{equation}

\begin{algorithm}
\caption{\textcolor{red}{Algorithm of the proposed LSSF-Net}}
\begin{algorithmic}[1]

\STATE \textbf{Input:} Input Image
\STATE \textbf{Output:} Segmented Output Image

\STATE Initialize parameters: filters, kernel sizes, pooling sizes, upsampling scales, etc.

\FOR{each convolutional block $i$}
    \STATE $\text{Conv}_i \gets \text{Convolution}(\text{Input}, \text{filters}_i, \text{kernel\_size}_i)$
    \STATE $\text{BN}_i \gets \text{BatchNormalization}(\text{Conv}_i)$
    \STATE $\text{ReLU}_i \gets \text{ReLU}(\text{BN}_i)$
    \IF{block has max pooling}
        \STATE $\text{Pooled}_i \gets \text{MaxPooling}(\text{ReLU}_i, \text{pool\_size}_i)$
    \ELSE
        \STATE $\text{Pooled}_i \gets \text{ReLU}_i$
    \ENDIF
    \STATE $\text{Input} \gets \text{Pooled}_i$
\ENDFOR

\IF{use GSASAB Layer}
    \STATE $\text{GSASAB\_out} \gets \text{GSASABLayer}(\text{Input})$
    \STATE $\text{Input} \gets \text{GSASAB\_out}$
\ENDIF

\IF{use Channel Shuffle}
    \STATE $\text{Shuffled} \gets \text{ChannelShuffle}(\text{Input})$
    \STATE $\text{Input} \gets \text{Shuffled}$
\ENDIF

\FOR{each upsampling block $j$}
    \STATE $\text{Upsample}_j \gets \text{Upsampling}(\text{Input}, \text{scale}_j)$
    \STATE $\text{Concat}_j \gets \text{Concatenate}(\text{Upsample}_j, \text{Feature\_Map}_j)$
    \IF{additional convolution is required}
        \STATE $\text{Conv}_j \gets \text{Convolution}(\text{Concat}_j, \text{filters}_j, \text{kernel\_size}_j)$
        \STATE $\text{Input} \gets \text{Conv}_j$
    \ELSE
        \STATE $\text{Input} \gets \text{Concat}_j$
    \ENDIF
\ENDFOR

\STATE $\text{Sigmoid\_out} \gets \text{Sigmoid}(\text{Input})$
\STATE $\text{Output} \gets \text{DicePixelClassificationLayer}(\text{Sigmoid\_out})$

\STATE \textbf{Return} Output

\end{algorithmic}
\end{algorithm}

Similarly, the output of the initial encoder block denoted by ($E_{o}$) is computed as (Eq. \ref{Eq3}).

\begin{equation}
E_{o}=m_{p}\left ( l^{3\times 3}\left ( l^{3\times 3}\left (s_{o}  \right ) \right ) \right )
\label{Eq3}
\end{equation}

where ($m_{p}$) is the maxpooling operation. The output of the encoder block $k^{th}$ ($E_{k}$) is computed by (Eq. \ref{Eq4}).

\begin{equation}
    E_{k}=m_{p}\left [ \Re\left \{ \beta _{n}\left ( f^{3\times 3}\left ( \beta _{n}\left ( f^{3\times 3}\left ( s_{k} \right ) \right ) \right ) \right ) + f^{3\times 3}\left ( l^{3\times 3}\left ( l^{3\times 3} \left ( E_{k-1} \right )\right ) \right )\right \} \right ]
    \label{Eq4}
\end{equation}

where ($s_{k}$) is the $k^{th}$ skip connection and is computed as given in (Eq. \ref{Eq5}).
\begin{equation}
    s_{k}=l^{3\times 3} (E_{k-1})
    \label{Eq5}
\end{equation}

Once the information is extracted by the encoder block, it is further refined by two consecutive attention blocks, named Self-Attention Block (SAB), to capture the contextual information from relative positions, followed by a Global Spatial Attention (GSA) block which is responsible for enhancing the local contextual information from a broader view through aggregating with global spatial information. In addition, we implemented a technique that involves channel splitting and shuffling to enhance the capabilities and efficiency of the LSSF-Net model. Channel splitting enables simultaneous processing of distinct channel subsets, promoting parallelisation. Concurrently, the technique of channel shuffling stimulates inter-channel interaction, thereby improving the overall information flow. Once the extracted feature information is further enhanced and refined, it is given to the decoder stage to reconstruct the spatial feature maps. Let ($D_{o}$) be the input given to the $k^{th}$ decoder block computed by (Eq. \ref{Eq6}).

\begin{equation}
    D_{o}=\texttt{GSAB}(E_{k}) \copyright \texttt{SAB}(E_{k})
    \label{Eq6}
\end{equation}

where $\copyright$ is the concatenation operation. To fuse the extracted feature information at the decoder stage, we have employed a conformer-based Focal Modulation Attention (CFMA) on the skip connections and then added this information by applying the ($l^{3\times 3}$) operation on the input coming from the $k^{th}$ decoder block and computed as (Eq. \ref{Eq7}).

\begin{equation}
    \Im _{k} = \texttt{CFMA}(s_{k}) + l^{3\times 3}(u_{p}(D_{k-1}))
    \label{Eq7}
\end{equation}
where $u_{p}$ is the upsampling operation that increases the spatial dimensions of the feature maps. The output of the $k^{th}$ decoder block is computed using (Eq. \ref{Eq8}).

\begin{equation}
    D_{k}=\Re \left [ f^{3\times 3}\left ( l^{3\times 3}\left ( l^{3\times 3}\left ( u_{p}\left ( D_{k-1} \right ) \right ) \right ) \right )+\beta _{n}\left ( f^{3\times 3}\left ( \beta _{n}\left ( f^{3\times 3}\left ( \Im _{k} \right ) \right ) \right ) \right ) \right ]
    \label{Eq8}
\end{equation}

The output of the model ($X_{out}$) is computed by applying the $l^{3\times 3}$ operation followed by the ($f^{1\times 1}$) convolution and the sigmoid ($\sigma$) operation as shown in (Eq. \ref{Eq9}).

\begin{equation}
    X_{out}= \sigma(f^{1\times 1}(l^{3\times}(\Im _{k})))
    \label{Eq9}
\end{equation}

The final binary predicted mask of size $256\times 256$ is obtained by employing the dice pixel classification layer on the model output.

\begin{figure}
    \centering
    \includegraphics[width=\columnwidth]{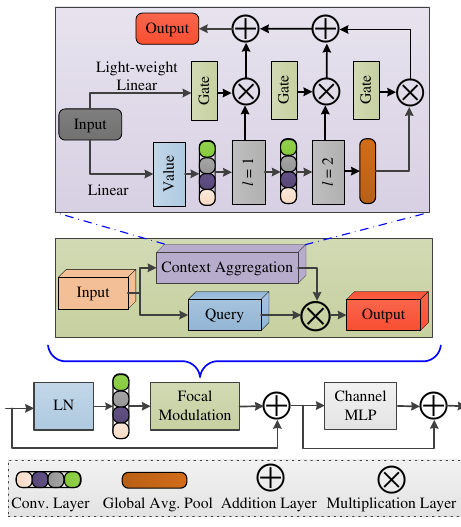}
    \caption{Schematic of the Conformer-based Focal Modulation
 Attention (CFMA), ``LN'' is the layer normalization.}
    \label{fig:CFMA}
\end{figure}

\subsection{Conformer-based Focal Modulation Block }

The conformer-based focal modulation block (CFMA) is introduced in the skip connections of the proposed LSSF-Net to further capture multiscale global semantic features, as shown in Fig. \ref{fig:CFMA}. The CFMA block takes the input ($C_{in}$) from the encoder block and applies the layer normalisation ($\texttt{LN}$) operation, followed by the $3\times 3$ convolution operation ($f^{3\times 3}$) and the focal modulation block ($\texttt{FMB}$) and adds the ($In$) with it as shown in Eq.\ref{Eq:CFMA1}.

\begin{equation}
C_{1}= In + \texttt{LN}(f^{3\times 3}(\texttt{FMB}(C_{in})))
    \label{Eq:CFMA1}
\end{equation}

The $\texttt{FMB}$ is a key component of the CFMA block and is designed to produce different scales of receptive fields in an adaptive manner. This is achieved by employing a contextual aggregation block to capture information at various scales, enabling the network to gather rich semantic information from the input data. The output ($C_{out}$) of CFMA is computed by applying the multilayer perception (MLP) of the channel to ($C_{1}$) as shown in the equation. \ref{Eq:CFMA2}.

\begin{equation}
C_{out}= C_{1} + \texttt{MLP}(C_{1})
    \label{Eq:CFMA2}
\end{equation}

Incorporating residual connections into the CMFA is essential to prevent the vanishing gradient issue during training. These connections enable gradients to pass directly through the block, enhancing the integration of more complex features across various scales. This approach enhances gradient flow and aids in training deeper networks, thereby simplifying the process of learning valuable data representations.

\begin{algorithm}
\caption{\textcolor{red}{Self-aware Attention Block}}
\label{alog_1}
\begin{algorithmic}[1]
\STATE \textbf{Input:} $x$ (input tensor), $temperature$, $dropout$
\STATE \textbf{Output:} Output tensor with scaled dot-product attention
\STATE Initialize $temperature \gets \sqrt{temperature}$ and $dropout$

\STATE Extract $B, H, W, C$ from the shape of $x$
\STATE Reshape $x$ to get $query$, $key$, and $value$
\STATE Permute dimensions of $query$

\STATE Calculate $energy \gets matmul(query, key)$
\STATE Divide $energy$ by $temperature$
\STATE Calculate $attention \gets softmax(energy)$
\STATE Apply dropout to $attention$

\STATE Calculate $output \gets matmul(value, attention)$
\STATE Reshape $output$ back to original input dimensions

\RETURN $output$
\end{algorithmic}
\end{algorithm}

\subsection{Self-aware Attention Block }
Self-aware Attention Block (SAB) is a type of multihead attention that can learn self-correlation but lacks the ability to learn spatial information; a commonly used approach in academic work is to pass the feature map to a position encoding block and then input it into the multihead attention block, as shown in algorithm \ref{alog_1}. The input feature map $F_{in}$ is then embedded in three matrices $Q\in R^{(h\times w)\times c}, K\in R^{C\times (h\times w)}, V\in R^{c\times (h\times w)}$,
\begin{equation}
    Q =  W_Q\centerdot F_{in}
\end{equation}
\begin{equation}
    K =  W_K\centerdot F_{in}
\end{equation}
\begin{equation}
    V =  W_V\centerdot F_{in}
\end{equation}
where $W_Q, W_K, W_V$ are three embedding functions for different linear projections. The scaling of the operation of the dot product with Softmax normalisation between $Q$ and $K$ gives $S\in R^{c\times c}$, which represents the similarity between the channels in $Q$ and others. To derive the aggregated values weighted by attention weights, the contextual attention map $S_{ct}$ is applied to the value matrix $V$. This process can be expressed through the multi-head attention mechanism, formulated as follows: 
\begin{equation}
    A_{tsa}(Q, K, V) = Softmax\frac{QK}{\sqrt{d_k}}V
\end{equation}
Finally, the $A_{tsa} \in R^{c\times (h\times w)}$ is reshaped to $R^{h\times w \times c}$, which is as the same as the input shape.

\begin{algorithm}
\caption{\textcolor{red}{Global Spatial Attention}}
\label{alog_2}
\begin{algorithmic}[1]
\STATE \textbf{Input:} $x$ (input tensor), $in\_channel$, $factor$
\STATE \textbf{Output:} Output tensor with global spatial attention
\STATE Initialize $in\_channel$ and $factor$
\STATE Calculate $dim \gets H \times W$ from input shape
\STATE Initialize trainable weight matrix $W \in \mathbb{R}^{dim \times dim}$ with random normal distribution

\STATE Extract $B, H, W, C$ from the shape of $x$
\STATE Apply $1 \times 1$ convolution on $x$ to get $proj\_query$ with reduced filters by $factor$
\STATE Reshape $proj\_query$ to $(B, H \times W, -1)$

\STATE Apply $1 \times 1$ convolution on $x$ to get $proj\_key$
\STATE Reshape $proj\_key$ to $(B, H \times W, -1)$ and permute dimensions

\STATE Calculate $energy \gets matmul(proj\_query, proj\_key)$
\STATE Calculate $attention \gets softmax(energy)$

\STATE Apply $1 \times 1$ convolution on $x$ to get $proj\_value$
\STATE Reshape and permute $proj\_value$
\STATE Calculate $output \gets matmul(proj\_value, attention)$
\STATE Multiply $output$ with the weight matrix $W$
\STATE Reshape and permute $output$ back to original input dimensions

\RETURN $output + x$
\end{algorithmic}
\end{algorithm}

\subsection{Global Spatial Attention }

Global spatial attention (GSA) is used to capture information on global position dependencies, as shown in algorithm \ref{alog_2}. The input feature map $F_{in}\in R^{h\times w\times c}$ is first embedded in $F^c\in R^{h\times w\times c}$ and $F^{c'}\in R^{h\times w\times c'}$ where $c' = c/2$. The reshape $F^{c'}\in R^{h\times w\times c'}$ to $F^{c'}_1\in R^{(h\times w)\times c'}$ and $F^{c'}_2\in R^{c'\times (h\times w)}$, respectively, the scaled dot product of $F^{c'}_1$ and $F^{c'}_2$then passes to a Softmax normalisation layer, the output map $S\in R^{(h\times w)\times (h\times w)}$ indicates spatial similarity, where $S_{i,j}$ represents the correlations between position $i^{th}$ and position $j^{th}$. The multi-head attention mechanism can be represented as
\begin{equation}
    A_{gsa} = Softmax(F^{c'}_1\cdot F^{c'}_2)F^c 
            = \frac{f^{c'}_1\cdot f^{c'}_2}{F^c}
\end{equation}

\subsection{Split Chanel-Shuffle }

Channel Shuffle is a technique that improves the flow of information across feature channels in a convolutional neural (CN) network. In group convolution, where input data from different groups is processed separately, the input and output channels are typically isolated. To overcome this, Channel Shuffle rearranges the channels by dividing them into subgroups. These subgroups are then mixed and fed into different groups in the next layer, ensuring that all channels can interact and share information effectively. This enhances the network's ability to learn from diverse features.

This process is carried out efficiently and seamlessly using a channel shuffle operation. A convolutional neural layer with $g$ groups and $n$ output channels, the output channels are first reshaped into dimensions of $(g, n / g)$, then transposed, and finally flattened back into a single dimension to serve as input for the next layer. Additionally, incorporating a split operation can make the model lighter by dividing the feature maps into smaller parts for more efficient processing. Split Channel Shuffle (SCS) is also differentiable and model-lightening, enabling its integration into network structures for end-to-end training.

\begin{equation}
\text{Output} \in {R}^{H \times W \times n} \rightarrow {R}^{H \times W \times g \times \frac{n}{g}}
\end{equation}
\begin{equation}
\text{Transpose}({R}^{H \times W \times g \times \frac{n}{g}}) \rightarrow {R}^{H \times W \times \frac{n}{g} \times g}
\end{equation}
\begin{equation}
\text{Flatten}({R}^{H \times W \times \frac{n}{g} \times g}) \rightarrow {R}^{H \times W \times n}
\end{equation}

\begin{table}
  \centering
  \caption{Description of the skin lesion segmentation datasets used for experimentation and evaluation of the proposed LSSF-Net.}
     \adjustbox {max width=0.5\textwidth}{%
 \begin{tabular}{lcccc}
\toprule
\multicolumn{1}{l}{\multirow{2}{*}{\textbf{Dataset}}}&\multicolumn{3}{c}{\textbf{Number of Images}}&\multicolumn{1}{c}{\multirow{2}{*}{\textbf{Resolution}}}\\
\cmidrule{2-4}
& \textbf{Train} & \textbf{Validation} & \textbf{Test} & \\
\toprule
ISIC2016 \cite{gutman2016skin} & 900 & N.A & 379 & 679$\times$453 - 6748$\times$4499\\
ISIC2017 \cite{codella2018skin} & 2000 & N.A & 600 & 679$\times$453 - 6748$\times$4499\\
ISIC2018 \cite{codella2019skin} & 2594& 100& 1000& 679$\times$453 - 6748$\times$4499\\
PH2 \cite{mendoncca2013ph} & 200 & N.A & N.A & 768$\times$560 \\
DDTI \cite{DDTIdataset} & 637 & N.A & N.A &$245\times 360$ - $560\times 360$\\
BUSI \cite{BUSIdataset} & 780 & N.A & N.A & $500\times500$ \\

\toprule
\end{tabular}
}
  \label{datasets}%
\end{table}%

\begin{figure*}
    \centering
    \includegraphics[width=\textwidth]{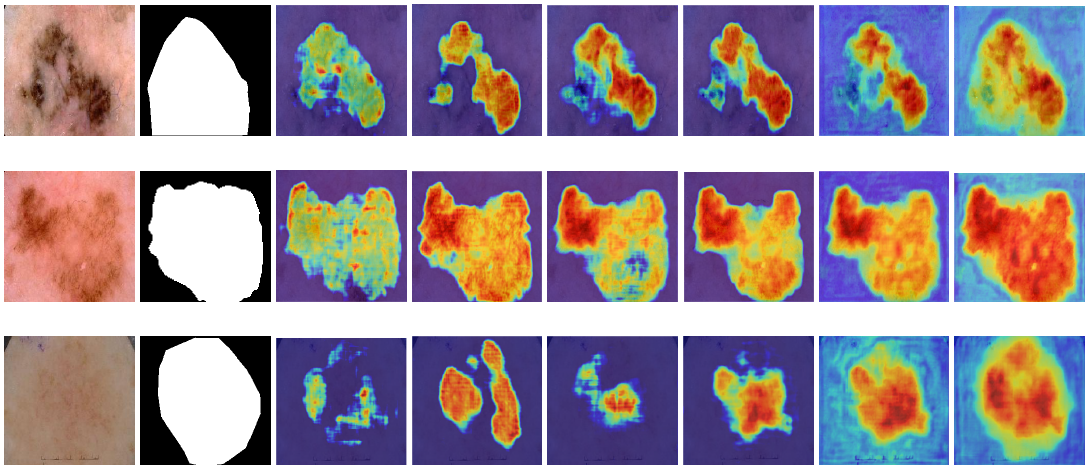}
    \caption{Visual results of ablation study on ISIC 2017 dataset. $1^{st}$ column shows the color image, $2^{nd}$ column shows the corresponding ground truth, $3^{rd}$ column shows the output of baseline network (BN), $4^{th}$ column shows the output of (BN + CFMA), $5^{th}$ column shows the output of (BN + SAB), $6^{th}$ column shows the output of (BN + CFMA + SAB), $7^{th}$ column shows the output of (BN + CFMA + SCS-SAB), and $8^{th}$ column shows the output of (BN + CFMA + SCS-SAB + Transfer Learning).}
    \label{fig:Vis_Ablation}
\end{figure*}

\begin{table}[htbp]
  \centering
  \caption{Performance based Ablation study of LSSF-Net on ISIC 2017 dataset. The "$\uparrow$" shows that the higher values are better.}
      \adjustbox {max width=0.5\textwidth}{%
    \begin{tabular}{lccccc}
    \toprule
    \multirow{2}[4]{*}{\textbf{Method}} & \multicolumn{5}{c}{\textbf{Performance Measures in (\%)}} \\
    \cmidrule{2-6}         &  $J_{ind}\uparrow$ & $D_{s}\uparrow$  & $A{cc}\uparrow$ & $S_{n}\uparrow$ & $S_{p}\uparrow$ \\
    \midrule
    Baseline  Network (BN) & 78.59 & 86.33 & 93.85 & 86.30 & 91.54 \\
    BN + CFMA & 79.80 & 86.73 & 94.39 & 86.50 & 92.79 \\
    BN + SAB & 81.31 & 88.30 & 94.99 & 90.14 & 93.28 \\
    BN + CFMA + SAB & 84.54 & 90.59 & 95.88 & 90.47 & 94.76 \\
    BN + CFMA + SCS-SAB & 85.27 & 91.14 & 96.07 & 91.20 & 94.98 \\
    BN + CFMA + SCS-SAB + Transfer Learning & 88.10 & 93.20 & 97.13 & 93.17 & 96.76 \\
    \bottomrule
    \end{tabular}%
    }
  \label{tab:ablation}%
\end{table}%

\section{Experiments and Results}
In this section, we will begin by providing a concise overview of the benchmark datasets used for skin lesion segmentation before delving into the experimental work of the proposed LSSF-Net.

\subsection{Datasets}
The effectiveness of the proposed LSSF-Net was assessed using four publicly available skin lesion datasets: three from the International Skin Imaging Collaboration (ISIC) archive and one from the PH2 dataset. Additionally, the model was evaluated on two ultrasound image datasets to further validate its performance. A detailed description of these datasets is provided below, and their distribution is presented in Table~\ref{datasets}.

\textbf{ISIC 2016:} The ISIC 2016 \cite{gutman2016skin} dataset includes 900 dermoscopic images for training and 379 images for testing, each provided with corresponding ground truth masks.

\textbf{ISIC 2017:} The ISIC 2017 \cite{codella2018skin} dataset consists of a total of 2000 dermoscopic images accompanied by the corresponding ground truth masks. These images are allocated for training purposes. Furthermore, the data set includes 150 images for validation and an additional 600 images specifically designed to evaluate the performance of the developed framework.

\textbf{ISIC 2018:} The ISIC 2018 \cite{tschandl2018ham10000,codella2019skin} dataset comprises 2594 dermoscopic images accompanied by their corresponding ground truth masks, which are used for training purposes. Additionally, the dataset includes 1000 images specifically designated for testing.

\textbf{PH2:} The PH2 \cite{mendoncca2013ph} dataset is a collection of 200 dermoscopic images accompanied by ground truth masks.


\textbf{DDTI:}The DDTI dataset \cite{DDTIdataset} consist of 637 ultrasound thyroid nodule images stored in the PNG format. These images show various resolutions, including $560\times 360$, $280\times 360$, and $245\times 360$ pixels. To ensure uniformity in image dimensions, all images are resized to $256\times 256$ pixels. The dataset is partitioned into training, validation, and test sets with proportions of 80\%, 10\%, and 10\%, respectively. In addition, performance evaluation employs a three-fold cross-validation approach.

\textbf{BUSI:} The BUSI dataset \cite{BUSIdataset} is composed of 780 breast ultrasound images obtained from women between 25 and 75 years of age. These images, available in PNG format, exhibit an average size of $500\times500$ pixels. Ground truth images, classified into three classes (normal, benign, and malignant), are provided for all instances. For consistency in image sizes, a uniform resizing of $256\times 256$ pixels is applied. The dataset is stratified into training, validation, and test sets, following the distribution of 80\%, 10\%, and 10\%, respectively. In addition, a three-fold cross-validation methodology is adopted for performance assessment.
 
\begin{table}
  \centering
  \caption{Computational Complexity Analysis of the Ablation Study for the LSSF-Net.}
 \adjustbox {max width=0.5\textwidth}{%
    \begin{tabular}{lcccc}
    \toprule
    \multirow{2}[4]{*}{\textbf{Method}} & \multicolumn{3}{c}{\textbf{Computational Analysis}} \\
\cmidrule{2-5}          & \textbf{Param (M) $\uparrow$} & \textbf{FLOPs (G) $\uparrow$} & \textbf{Inference Time (ms) $\uparrow$} & \textbf{ Jaccard $\downarrow$} \\
    \midrule
    Baseline  Network (BN)                  & \textbf{0.550}  & \textbf{2.57} & \textbf{5}    &  78.59 \\
    BN + CFMA                               & 0.745  & 3.10  & 10  &  79.80 \\
    BN + SAB                                & 0.616  & 2.57 & 6.7  &  81.31 \\
    \textbf{BN + CFMA + SCS-SAB}                     & 0.811  & 3.10 & 13.7 &  \textbf{85.27}\\
    BN + CFMA + SAB                         & 0.812  & 3.10 & 16.7 &  84.54\\
    \bottomrule
    \end{tabular}%
    }
  \label{tab:Computations_Abl}%
\end{table}%
\subsection{Performance Measures}
The proposed LSSF-Net's performance is assessed using five key metrics endorsed by the ISIC challenge leaderboard: accuracy, Jaccard index (IOU), Dice coefficient, sensitivity, and specificity. These metrics are determined based on the counts of true positives (TP), true negatives (TN), false positives (FP), and false negatives (FN), as outlined in equations (7-11).

\begin{equation}\tag{7}
\mathrm{Accuracy (A_{cc})} = \frac{{T_{P}+T_{N}}}{{T_{P}+T_{N}+F_{P}+F_{N}}}
\end{equation}
\begin{equation}\tag{8}
\mathrm{Sensitivity (S_{n})} = \frac{{T_{P}}}{{T_{P} + F_{N}}}
\end{equation}
\begin{equation}\tag{9}
\mathrm{Jacard-Index (J_{ind})} = \frac{{T_{P}}}{{T_{P}+ F_{P} +F_{N}}}
\end{equation}
\begin{equation}\tag{10}
\mathrm{Dice-Score (D_{s})} = \frac{{2*T_{P}}}{{2*T_{P}+ F_{P} +F_{N}}}
\end{equation}
\begin{equation}\tag{11}
\mathrm{Specificity (S_{p})} = \frac{{T_{N}}}{{T_{N} + F_{P}}}
\end{equation}

\subsection{Implementation Details}
Initially, all training images are reshaped to $256\times 256$ and then fed to the LSSF-Net. Adam is employed as the optimiser with $\beta_1 = 0.90, \beta_2 = 0.999$, where $\beta_1, \beta_2$ are the initial decay rates adopted when estimating the first and second moments of the gradient that are multiplied at the end of each epoch. The adoption of the values is based on the study of \cite{Mirikharaji2022} that $\beta_1 = 0.90, \beta_2 = 0.999$ are the values most commonly used in previous articles on the analysis of skin lesions. Similarly, based on the statistical results of the literature, the initial learning rate is set to 0.001. Meanwhile, the Early Stop monitor is set and starts from the $10^{th}$ epoch to terminate the training process if the monitored metric does not improve for 9 epochs. It is worth mentioning that due to the dynamic components of the proposed loss function, the model converged to the lower-loss direction but the Jaccard index is also decreased in some cases, which reflects the proposed loss being highly sensitive to the dynamic weights. To improve the adaptability of the model on different data sets, once $L_{bl}$ is involved, the Jaccard coefficient is manually adopted as the training monitor and validation loss value otherwise. All experiments are performed on a local PC with a GPU NVIDIA GeForce RTX 3090 with batch size 24 on the Keras framework with Python 3.9.

\begin{table}
  \centering
  \caption{Performance comparison of the proposed LSSF-Net with state-of-art on ISIC 2018 Dataset. The best scores are presented in \textbf{bold}.}
          \adjustbox {max width=0.5\textwidth}{%
    \begin{tabular}{lccccc}
    \toprule
    \multirow{2}[4]{*}{\textbf{Method}} & \multicolumn{5}{c}{\textbf{Performance Measures in (\%)}} \\
\cmidrule{2-6}          & $J_{ind}\uparrow$ & $D_{s}\uparrow$  & $A{cc}\uparrow$ & $S_{n}\uparrow$ & $S_{p}\uparrow$ \\
    \midrule
    U-Net \cite{ronneberger2015u}  & 80.09 & 86.64 & 92.52 & 85.22 & 92.09 \\
    BCDU-Net \cite{azad2019bi} & 81.10 & 85.10 & 93.70 & 78.50 & 97.20 \\
    DAGAN \cite{Lei2020}  & 81.13 & 88.07 & 93.24 & 90.72 & 95.88 \\
    UNet++ \cite{Zhou2018} & 81.62 & 87.32 & 93.72 & 88.70 & 93.96 \\
    FAT-Net \cite{Wu2021FAT} & 82.02 & 89.03 & 95.78 & 91.00 & 96.99 \\
    Swin-Unet \cite{cao2023swin} & 82.79 & 88.98 & 96.83 & 90.10 & 97.16 \\
    FTN Network \cite{he2022fully} & 82.80 & 89.80 & 96.20 & 96.20 & 97.50 \\
    AS-Net \cite{HU2022117112}  & 83.09 & 89.55 & 95.68 & 93.06 & 94.69 \\
    DCSAU-Net \cite{xu2023dcsau} & 83.10 & 89.40 & 95.86 & 91.09 & - \\
    ICL-Net \cite{cao2022icl}   & 83.76 & 90.41 & 97.24 & 91.66 & 97.63 \\
    Ms RED \cite{Dai2022} & 83.86 & 90.33 & 96.45 & 91.10 & - \\
    DconnNet \cite{yang2023directional} & 83.91 & 90.43 & 96.39 & -     & - \\
    ARU-GD \cite{Maji2022} & 84.55 & 89.16 & 94.23 & 91.42 & 96.81 \\
    \midrule
    \textbf{Proposed LSSF-Net} & \textbf{89.06} & \textbf{93.77} & 96.43 & \textbf{94.33} & 93.18 \\
    \bottomrule
    \end{tabular}%
    }
  \label{tab:ISIC2018}%
\end{table}%

\begin{figure*}
    \centering
    \includegraphics[width=\textwidth]{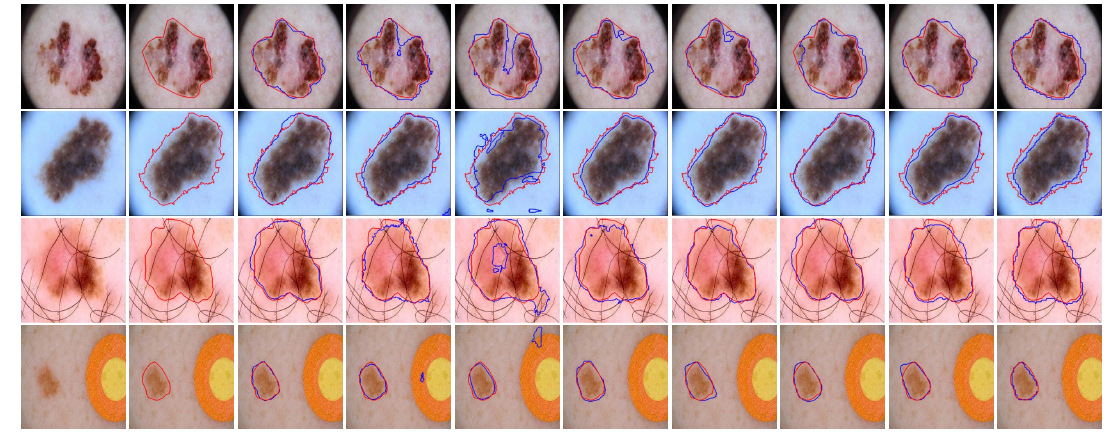}
     \caption{Comparison of the visual performance of the proposed LSSF-Net on ISIC 2018 \cite{codella2019skin} dataset.}
    \label{fig:Vis_ISIC2018}
\end{figure*}

\subsection{Loss Function}
In this paper, we used a combined loss consisting of Binary Cross Entropy and Jaccard losses to guide the training process. Denote $G$ as the ground truth set and $P$ as the model prediction map. $p_ic$ indicates the probability that the pixel $i$ belongs to the class $c$, $g_ic$ indicates the ground truth label. $\epsilon$ in the following representations is the smooth index.

\subsubsection{Binary Cross Entropy}
Cross-entropy quantifies the divergence between two probability distributions. In the context of binary segmentation, the binary cross-entropy loss function is expressed as
\begin{align*}
    L_{bce} = -\sum_{i=1}^{N}g_{ic}logp_{ic} + (1-g_{ic})log(1-p_{ic})\tag{12}
\end{align*}

\subsubsection{Jaccard Loss} The Jaccard coefficient is an index that assesses the similarity between the ground truth and segmentation sets by calculating the ratio of the intersection over the union, where
\begin{align*}
    IoU_c &= \frac{\vert G\cap P\vert}{\vert G\cup P\vert} = \frac{\vert G\cap P\vert}{\vert G\vert + \vert P\vert - \vert G \cap P\vert}\\
    &= \frac{\sum_{i=1}^Np_{ic}g_{ic}+\epsilon}{\sum^N_{i=1}p_{ic}+g_{ic}-p_{ic}g_{ic}+\epsilon}\tag{13}
\end{align*}
The Jaccard coefficient loss $L_{jcd}$ is defined as the minimization of $IoU_c$, where
\begin{align*}
    L_{jcd} = \sum_c 1- IoU_c\tag{14}
\end{align*}

\begin{figure*}
    \centering
    \includegraphics[width=\textwidth]{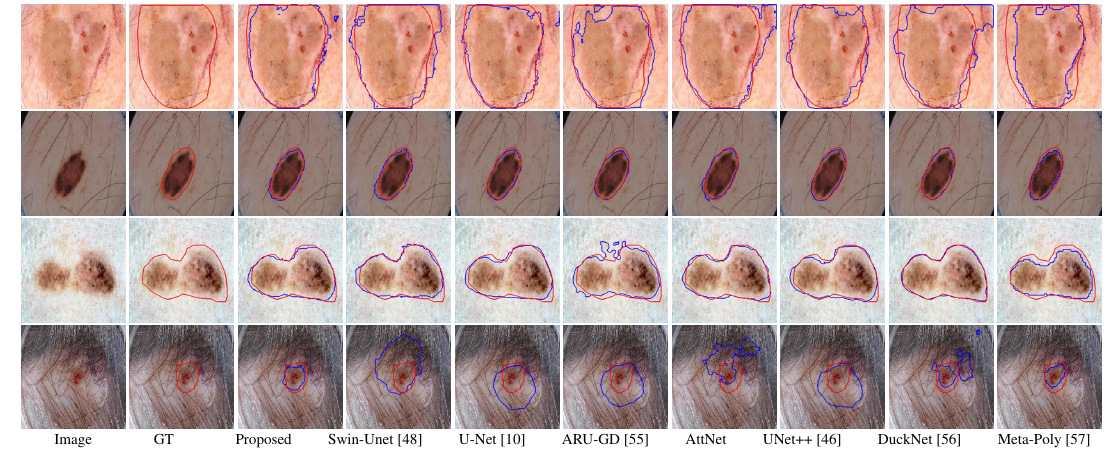}
 \caption{Comparison of the visual performance of the proposed LSSF-Net on ISIC 2017 \cite{codella2018skin} dataset.}
    \label{fig:Vis_ISIC2017}
\end{figure*}

\subsection{Ablation Study of LSSF-Net on ISI2017 Dataset}
The ablation study for the LSSF-Net on the ISIC 2017 dataset aims to evaluate the impact of different network components and strategies on the model's performance. By systematically adding and modifying various modules within the network, we can determine their individual and combined contributions to the overall efficacy of the LSSF-Net. This study provides insights into how each component improves the network's ability to accurately segment skin lesions, thereby informing future improvements and optimisations. The experiments for LSSF-Net are extensively conducted using the ISIC-2017 dataset. Table \ref{tab:ablation} presents the quantitative improvements achieved by the proposed LSSF-Net. The ablation study begins by implementing a basic UNet-based CNN model with booster connection, which serves as the baseline for comparison. After that, conformer focal modulation attention (CFMA) is employed in the skip connections. The second experiment is carried out employing a self-attention block (SAB) in the bottleneck of the network. In the third experiment, both CFMA and SAB are incorporated. It should be mentioned that this combination has significantly improved overall performance. After that, split-channel-shuffle-based SAB (SCS-SAB) is employed in the bottleneck layer of the network. Finally, in the last experiment, the transfer learning strategy is employed to take advantage of domain knowledge. 

The proposed LSSF-Net leverages pre-trained weights from the ISIC 2016, 2017, and 2018 datasets to enhance its performance on these datasets. Specifically, for transfer learning, we initialized the training of the ISIC 2016 and 2018 datasets with weights pre-trained on the ISIC 2017 dataset. In contrast, training in the ISIC 2017 dataset was initialised with weights pre-trained in the ISIC 2016 dataset. This approach of cross-dataset weight initialisation further improves the generalisation and performance of the model.

Figure \ref{fig:Vis_Ablation} presents the visual results of the ablation study in the ISIC 2017 dataset. The first column shows the RGB input image, the second column shows the corresponding ground truth images, and columns $3-8$ show the visual results of (BN + CFMA), (BN + SAB), (BN + CFMA + SAB), (BN + CFMA + SCS-SAB), and (BN + CFMA + SCS-SAB + Transfer Learning), respectively. It is evident from the figure \ref{fig:Vis_Ablation} that the performance of the proposed LSSF-Net is gradually enhanced by incorporating different modules into the baseline network. The computational complexity of the LSSF-Net shown in table \ref{tab:Computations_Abl}  and its variants is crucial for understanding their efficiency and feasibility for practical applications. In this study, we analyze the number of parameters (Param), floating point operations per second (FLOPs), and inference time for each model variant. The baseline network (BN) serves as a reference point, and we assess the impact of adding CFMA, SAB, and SCS-SAB modules on computational demands. The baseline model has the fewest parameters at 0.550 million and a Jaccard score of 78.59. However, as more complex modules like CFMA and SAB are added, the number of parameters and computational demands increase, but so does the performance. For instance, the combination of BN + CFMA + SCS-SAB achieves a higher Jaccard score of 85.27 with 0.811 million parameters. This analysis helps identify the trade-offs between model complexity and performance, guiding the selection of the most efficient network configuration for real-world deployment.
\begin{table}
  \centering
  \caption{Performance comparison of the proposed LSSF-Net with state-of-art on ISIC 2017 Dataset. The best scores are presented in \textbf{bold}.}
  \adjustbox {max width=0.5\textwidth}{
    \begin{tabular}{lccccc}
    \toprule
    \multirow{2}[4]{*}{\textbf{Method}} & \multicolumn{5}{c}{\textbf{Performance Measures in (\%)}} \\
    \cmidrule{2-6} & $J_{ind}\uparrow$ & $D_{s}\uparrow$ & $Acc\uparrow$ & $S_{n}\uparrow$ & $S_{p}\uparrow$ \\
    \midrule
    U-Net \cite{ronneberger2015u}  & 75.69 & 84.12 & 93.29 & 84.30 & 93.41 \\
    DAGAN \cite{Lei2020}  & 75.94 & 84.25 & 93.26 & 83.63 & 97.24 \\
    ReGANet & 76.40 & 85.60 & 93.60 & 84.20 & 95.00 \\
    FAT-Net \cite{Wu2021FAT} & 76.53 & 85.00 & 93.26 & 83.92 & \textbf{97.25} \\
    Ms RED \cite{Dai2022} & 78.55 & 86.48 & 94.10 & -     & - \\
    UNet++ \cite{Zhou2018} & 78.58 & 86.35 & 93.73 & 87.13 & 94.41 \\
    BCDU-Net \cite{azad2019bi} & 79.20 & 78.11 & 91.63 & 76.46 & 97.09 \\
    SEACU-Net \cite{jiang2022seacu} & 80.50 & 89.11 & 95.35 & -     & - \\
    AS-Net \cite{HU2022117112}  & 80.51 & 88.07 & 94.66 & 89.92 & 95.72 \\
    ARU-GD \cite{Maji2022} & 80.77 & 87.89 & 93.88 & 88.31 & 96.31 \\
    Swin-Unet \cite{cao2023swin} & 80.89 & 81.99 & 94.76 & 88.06 & 96.05 \\
    BA-Net \cite{wang2022boundary} & 81.00 & 88.10 & 94.60 & 89.70 & 96.60 \\
    \midrule
    \textbf{Proposed LSSF-Net} & \textbf{88.09} & \textbf{93.20} & \textbf{97.13} & \textbf{93.16} & 96.76 \\
    \bottomrule
    \end{tabular}
  }
  \label{tab:ISIC2017}
\end{table}

\subsection{Results and Discussions}

This section starts with a performance comparison of the proposed LSSF-Net with recent methods in the data sets ISIC 2018 \cite{codella2019skin}, ISIC 2017 \cite{codella2018skin}, ISIC 2016 \cite{gutman2016skin} and PH2 \cite{mendoncca2013ph}. Most of the comparisons presented in Tables \ref{tab:ISIC2018}-\ref{tab:PH2} have been taken from the articles cited in the literature. However, we reproduced the results of the methods used for visual comparisons. Finally, we have also demonstrated the generalisation of the proposed LSSF-Net on two datasets of ultrasound images: BUSI \cite{BUSIdataset} for segmentation of breast cancer lesion and DDTI \cite{DDTIdataset} for segmentation of thyroid nodules. This generalisation shows the adaptability of the proposed LSSF-Net to other medical image segmentation modalities.

\subsubsection{Performance Comparisons on the ISIC 2018 dataset}

We compare the proposed LSSF-Net with 13 other cutting-edge methods in the ISIC 2018 dataset to determine how well our proposed LSSF-Net works. U-Net \cite{ronneberger2015u}, BCDU-Net \cite{azad2019bi}, DAGAN \cite{Lei2020}, UNet++ \cite{Zhou2018}, FAT-Net \cite{Wu2021FAT}, Swin-Unet \cite{cao2023swin}, FTN Network \cite{he2022fully}, AS-Net \cite{HU2022117112}, DCSAU-Net \cite{xu2023dcsau}, ICL-Net \cite{cao2022icl}, Ms RED \cite{Dai2022}, DconnNet \cite{yang2023directional}, and ARU-GD \cite{Maji2022} are included for comparisons. It is important to mention that, in addition to U-Net, BCDU-Net, UNet++, Swin-Unet, and ARU-GD, all the results are taken from the cited papers. To ensure equitable comparisons, all comparisons were performed under identical computational settings and data augmentations. Table \ref{tab:ISIC2018} presents the statistical results for skin lesion segmentation in the ISIC 2018 dataset. The proposed LSSF-Net has outperformed all other methods presented in table \ref{tab:ISIC2018} in terms of the Jaccard index. Compared to the methods listed, LSSF--Net scored 4. 5\% ----8. 9\%, better in terms of the Jaccard index in the ISIC 2018 dataset. In addition, we have presented several examples of segmentation outcomes for visual comparisons. During our experiments, we carefully chose the five methods (U-Net, BCDU-Net, UNet++, ARU-GD, and Swin-Unet) for the visual analysis shown in Figure \ref{fig:Vis_ISIC2018}. Our observations indicate a consistent outperformance of LSSF-Net, yielding superior segmentation results, particularly in challenging scenarios. All of these methods are flawed because they do not use global contextual information well enough and cannot accurately predict skin lesions when there is occlusion and low contrast between pixels in the foreground and background.

\subsubsection{Performance Comparisons on the ISIC 2017 dataset}

In the context of the ISIC 2017 dataset, we performed a comparative analysis between our proposed LSSF-Net and 11 state-of-the-art methods. This assessment is carried out in identical computing environments and uniform data augmentations for a fair and equitable evaluation. U-Net \cite{ronneberger2015u}, DAGAN \cite{Lei2020}, FAT-Net \cite{Wu2021FAT}, Ms RED \cite{Dai2022}, UNet++ \cite{Zhou2018}, BCDU-Net \cite{azad2019bi}, SEACU-Net \cite{jiang2022seacu}, AS-Net \cite{HU2022117112}, ARU-GD \cite{Maji2022}, Swin-Unet \cite{cao2023swin}, and BA-Net \cite{wang2022boundary} are included for comparison. It is important to mention that, in addition to U-Net, BCDU-Net, UNet++, Swin-Unet, and ARU-GD, all the results are taken from the cited papers. The proposed LSSF-Net has outperformed all other methods by scoring 4.39\%--12.4\% better Jaccard index. Furthermore, it is evident from Table \ref{tab:ISIC2017} that LSSF-Net consistently exceeds other competing methodologies in most metrics. In addition, we have presented several examples of segmentation outcomes for visual comparisons. During our experiments, we carefully chose the five methods (U-Net, BCDU-Net, UNet++, ARU-GD and Swin-Unet) for the visual analysis shown in Figure \ref{fig:Vis_ISIC2017}. Our observations indicate a consistent outperformance of LSSF-Net, yielding superior segmentation results, particularly in challenging scenarios. Even when dealing with skin lesions characterised by diverse scales and irregular shapes, LSSF-Net consistently achieves the best segmentation results that closely align with the truth of the ground.

\begin{table}
  \centering
  \caption{Performance comparison of the proposed LSSF-Net with state-of-art on ISIC 2016 Dataset. The best scores are presented in \textbf{bold}.}
          \adjustbox {max width=0.5\textwidth}{%
    \begin{tabular}{lccccc}
    \toprule
    \multirow{2}[4]{*}{\textbf{Method}} & \multicolumn{5}{c}{\textbf{Performance Measures in (\%)}} \\
\cmidrule{2-6}          & $J_{ind}\uparrow$ & $D_{s}\uparrow$  & $A{cc}\uparrow$ & $S_{n}\uparrow$ & $S_{p}\uparrow$ \\
    \midrule
   
    U-Net \cite{ronneberger2015u} & 81.38 & 88.24 & 93.31 & 87.28 & 92.88 \\
    Superpixels and Hybrid Texture \cite{dos2022semi} & 82.43 & -     & 96.24 & 86.12 & 95.62 \\
    UNet++ \cite{Zhou2018} & 82.81 & 89.19 & 93.88 & 88.78 & 93.52 \\
    BCDU-Net \cite{azad2019bi} & 83.43 & 80.95 & 91.78 & 78.11 & 96.20 \\
    CPFNet \cite{9049412} & 83.81 & 90.23 & 95.09 & 92.11 & 95.91 \\
    DAGAN \cite{Lei2020} & 84.42 & 90.85 & 95.82 & 92.28 & 95.68 \\
    ARU-GD \cite{Maji2022} & 85.12 & 90.83 & 94.38 & 89.86 & 94.65 \\
    FAT-Net \cite{Wu2021FAT} & 85.30 & 91.59 & 96.04 & 92.59 & 96.02 \\
    Ms RED \cite{Dai2022} & 87.03 & 92.66 & 96.42 & -     & - \\
    Swin-Unet \cite{cao2023swin} & 87.60 & 88.94 & 96.00 & 92.27 & 95.79 \\
    Hyper-Fusion Net \cite{bi2022hyper} & 88.17 & -     & 96.64 & 94.22 & 96.45 \\
    \midrule
    \textbf{Proposed LSSF-Net} & \textbf{93.04} & \textbf{96.30} & \textbf{98.25} & \textbf{96.41} & \textbf{97.52} \\

    \bottomrule
    \end{tabular}%
  \label{tab:ISIC2016}%
  }
\end{table}%

\begin{figure*}
    \centering
    \includegraphics[width=\textwidth]{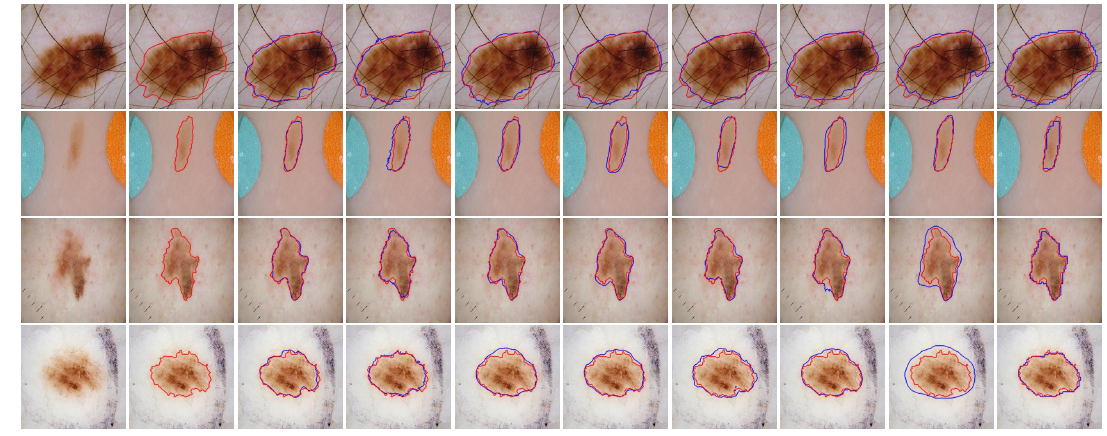}
    \caption{Comparison of the visual performance of the proposed LSSF-Net on ISIC 2016 \cite{gutman2016skin} dataset.}
    \label{fig:Vis_ISIC2016}
\end{figure*}
\subsubsection{Performance Comparisons on the ISIC 2016 dataset}

In the context of the ISIC 2016 dataset, we conducted a comparative analysis between our proposed LSSF-Net and ten state-of-the-art methods. This assessment is carried out under identical computing environments and uniform data augmentations for a fair and equitable evaluation. U-Net \cite{ronneberger2015u}, UNet++ \cite{Zhou2018}, BCDU-Net \cite{azad2019bi}, CPFNet \cite{9049412}, DAGAN \cite{Lei2020}, ARU-GD \cite{Maji2022}, FAT-Net \cite{Wu2021FAT}, Ms RED \cite{Dai2022}, Swin-Unet \cite{cao2023swin}, and Hyper-Fusion Net \cite{bi2022hyper} are included for comparison. It is important to mention that, in addition to U-Net, BCDU-Net, UNet++, Swin-Unet, and ARU-GD, all the results are taken from the cited papers. The proposed LSSF-Net has outperformed all other methods by scoring 4.87\%--11.66\% better Jaccard index. Furthermore, it is evident from Table \ref{tab:ISIC2016} that LSSF-Net consistently exceeds other competing methodologies in all metrics. Furthermore, we have presented several examples of segmentation outcomes for visual comparisons. During our experiments, we carefully chose the five methods (U-Net, BCDU-Net, UNet++, ARU-GD and Swin-Unet) for the visual analysis shown in Figure \ref{fig:Vis_ISIC2016}. Our observations consistently demonstrate the superior performance of LSSF-Net, especially evident in challenging scenarios, resulting in superior segmentation outcomes. Even when faced with skin lesions exhibiting diverse scales and irregular shapes, LSSF-Net consistently achieves optimal segmentation results.

\subsubsection{Performance Comparisons on the PH2 dataset}
\begin{figure*}
    \centering
    \includegraphics[width=\textwidth]{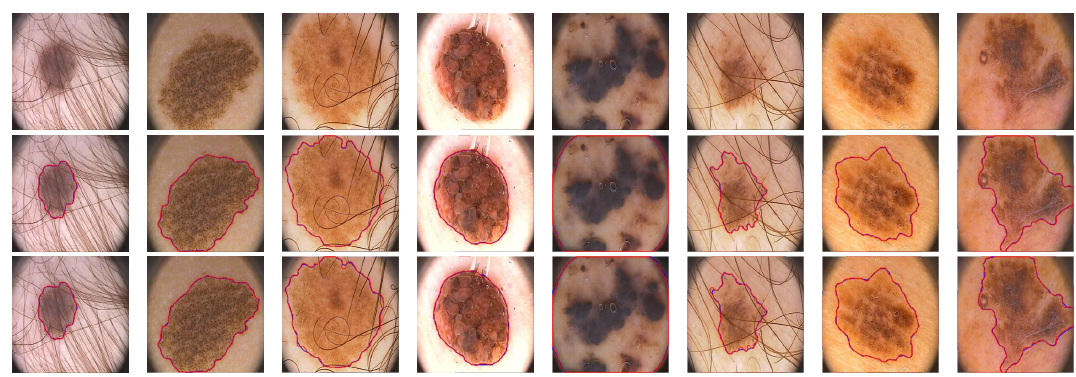}
\caption{Visual results of the proposed LSSF-Net on the PH2 \cite{mendoncca2013ph} dataset. The first row displays the RGB images, the second row shows the ground truth, and the third row presents the segmentation outputs from LSSF-Net, with training conducted on ISIC 2016 and testing on PH2.}
    \label{fig:Vis_PH2}
\end{figure*}

Finally, the generalisation of the proposed LSSF-Net is accessed with cross-dataset validation of the proposed LSSF-Net. The experimental results are calculated with training on ISIC 2016 and tested on the PH2 \cite{mendoncca2013ph} dataset. Performance of the proposed LSSF-Net in the PH2 \cite{mendoncca2013ph} dataset with various state-of-the-art methods, including MFCN \cite{bi2019step}, DCL-PSI \cite{bi2019step}, ICL-Net \cite{cao2022icl} and AS-Net \cite{HU2022117112}. Table \ref{tab:PH2} presents the performance comparison of the proposed LSSF-Net with the latest methods. Compared to state-of-the-art methods, the Jaccard index of the proposed LSSF-Net is improved by 3.91\%--7.72\% in the PH2 dataset \cite{mendoncca2013ph}. Figure \ref{fig:Vis_PH2} presents the visual results of LSSF-Net in the PH2 dataset. The first row shows the RGB input images, the second row shows the corresponding ground truth images, and the third column shows the output of the proposed LSSF-Net. It can be seen in Figure \ref{fig:Vis_PH2} that the proposed LSSF-Net accurately segments the lesion region in the presence of hair, contrast variations, variation in the size of the lesion, and irregular boundary shapes.

\subsubsection{Cross Dataset Performance Evaluation}
To demonstrate the robust generalization of the proposed LSSF-Net, cross-dataset evaluations have been conducted. Table \ref{tab:cross_dataset} presents the performance metrics of LSSF-Net across different datasets, where the model has been trained on one dataset and tested on others. The results indicate strong generalization capability. Specifically, the $J_{ind}$ score of LSSF-Net on the ISIC 2017 dataset has shown only a 2\% decrease when trained on the ISIC 2016 dataset and a 3\% decrease when trained on the ISIC 2018 dataset. Similarly, training on the ISIC 2017 dataset and testing on the ISIC 2016 and ISIC 2018 datasets has resulted in a 3\% and 4\% drop in $J_{ind}$, respectively. Lastly, training on the ISIC 2018 dataset and testing on the ISIC 2016 and ISIC 2017 datasets has yielded $J_{ind}$ reductions of 2.24\% and 3\%, respectively.
\begin{table}
  \centering
  \caption{Cross dataset validation of the proposed LSSF-Net.}
   \adjustbox {max width=0.5\textwidth}{%
    \begin{tabular}{llccccc}
    \toprule
    \multirow{2}[4]{*}{\textbf{Training Dataset}} & \multirow{2}[4]{*}{\textbf{Testing Dataset}} & \multicolumn{5}{c}{\textbf{Performance Measures in (\%)}} \\
\cmidrule{3-7}          &       &   $Jaccard$ & $Dice$  & $Acc$ & $Sen$ & $Sp$ \\
    \midrule
    \multirow{2}[2]{*}{\textbf{ISIC 2016}} & ISIC 2017 & 81.06 & 87.53 & 95.66 & 87.65 & 95.04 \\
          & ISIC 2018 & 88.29 & 93.13 & 95.98 & 93.75 & 92.87 \\
          & PH2 & 90.63 & 94.93 & 96.85 & 95.29 & 94.24 \\
    \midrule
    \multirow{2}[2]{*}{\textbf{ISIC 2017}} & ISIC 2016 & 92.65 & 96.26 & 98.25 & 96.63 & 97.55 \\
          & ISIC 2018 & 88.82 & 93.46 & 96.19 & 94.06 & 92.86 \\
          & PH2 & 91.13 & 95.36 & 97.11 & 95.81 & 93.60 \\
    \midrule
    \multirow{2}[2]{*}{\textbf{ISIC 2018}} & ISIC 2016 & 91.28 & 95.25 & 97.74 & 95.18 & 97.13 \\
          & ISIC 2017 & 85.65 & 91.30 & 96.68 & 91.26 & 96.32 \\
          & PH2 &90.58 & 94.94 & 96.72 & 95.21 & 93.83 \\
    \bottomrule
    \end{tabular}%
    }
  \label{tab:cross_dataset}%
\end{table}%

\begin{table}
  \centering
  \caption{Performance comparison of the proposed LSSF-Net with state-of-art on PH2 Dataset. The best scores are presented in \textbf{bold}.}
 \adjustbox {max width=0.5\textwidth}{%
  \begin{tabular}{lccccc}
    \hline
   \multirow{2}[4]{*}{\textbf{Method}} & \multicolumn{5}{c}{\textbf{Performance Measures in (\%)}} \\
\cmidrule{2-6}          & $J_{ind}\uparrow$ & $D_{s}\uparrow$  & $A{cc}\uparrow$ & $S_{n}\uparrow$ & $S_{p}\uparrow$ \\
   \hline
    MFCN \cite{bi2019step}  & 83.99 & 90.66 & 94.24 & 94.89 & 93.98 \\
    DCL-PSI \cite{bi2019step} & 85.90 & 92.10 & 95.30 & 96.23 & 94.52 \\
    ICL-Net \cite{cao2022icl}  & 87.25 & 92.80 & 96.32 & 95.46 & \textbf{97.36} \\
    AS-Net \cite{HU2022117112}   & 87.60 & 93.05 & 95.20 & \textbf{96.24} & 94.31 \\
\hline
    \textbf{Proposed LSSF-Net}  & \textbf{91.71} & \textbf{95.57} & \textbf{97.24} & 95.92 & 94.43 \\
    \hline
    \end{tabular}%
    }
  \label{tab:PH2}%
\end{table}

\begin{table}
  \centering
  \caption{Performance comparison of LSSF-Net model with various state-of-the-art methods on the breast lesion segmentation dataset BUSI.}
    \adjustbox{max width=\textwidth}{%
    \begin{tabular}{lccccc}
    \toprule
    \multirow{2}[4]{*}{\textbf{Method}} & \multicolumn{5}{c}{\textbf{Performance Measures in (\%)}} \\
    \cmidrule{2-6} & $J_{ind}\uparrow$ & $D_{s}\uparrow$  & $A{cc}\uparrow$ & $S_{n}\uparrow$ & $S_{p}\uparrow$ \\
    \midrule
    U-Net \cite{ronneberger2015u}  & 67.77 & 76.96 & 95.48 & 78.33 & 96.13 \\
    FPN \cite{lin2017feature} & 74.09 & 82.67 & -     & 85.39 & - \\
    DeeplabV3+ \cite{chen2018encoder} & 73.48 & 82.68 & -     & 83.37 & - \\
    ConvEDNet \cite{lei2018segmentation} & 73.57 & 82.70 & -     & 85.51 & - \\
    UNet++ \cite{zhou2018unet++} & 76.85 & 76.22 & 97.97 & 78.61 & 98.86 \\
    BCDU-Net \cite{azad2019bi} & 74.49 & 66.75 & 94.82 & 86.85 & 95.57 \\
    BGM-Net \cite{wu2021bgm} & 75.97 & 83.97 & -     & 83.45 & - \\
    ARU-GD \cite{Maji2022} & 77.07 & 83.64 & 97.94 & 83.80 & 98.78 \\
    Swin-Unet \cite{cao2023swin} & 77.16 & 84.45 & 97.55 & 84.81 & 98.34 \\
    \midrule
    \textbf{LSSF-Net} & \textbf{92.99} & \textbf{96.34} & \textbf{99.55} & \textbf{96.58} & \textbf{99.72} \\
    \bottomrule
    \end{tabular}%
  }
  \label{tab:BUSI}
\end{table}

    \begin{figure*}
    \centering
    \includegraphics[width=\textwidth]{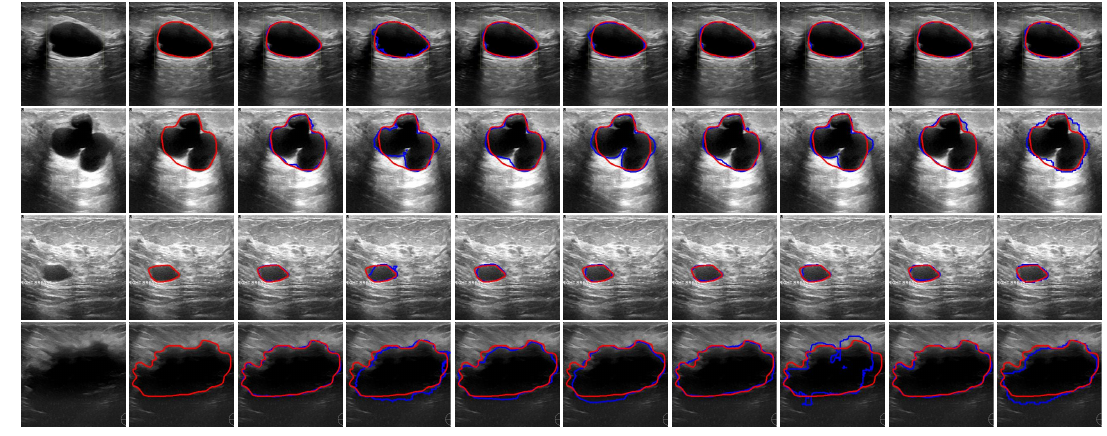}
 \caption{Comparison of the visual performance of the proposed LSSF-Net on BUSI  \cite{BUSIdataset} dataset.}
    \label{fig:Vis_BUSI}
\end{figure*}
\subsubsection{Generalisation of the Proposed LSSF-Net}

\begin{figure*}
    \centering
    \includegraphics[width=\textwidth]{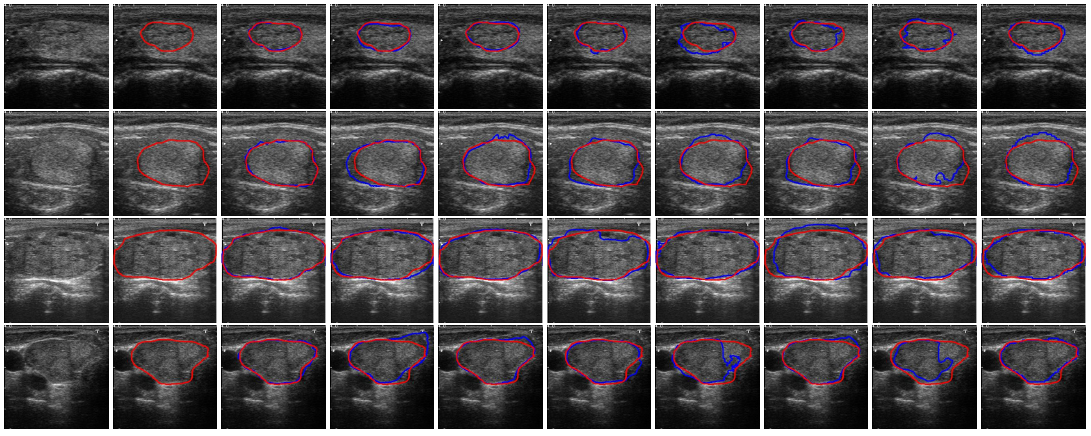}
  \caption{Comparison of the visual performance of the proposed LSSF-Net on DDTI  \cite{DDTIdataset} dataset.}
    \label{fig:Vis_DDTI}
\end{figure*}

The efficacy of LSSF-Net for thyroid nodule image segmentation has been assessed using the publicly accessible DDTI \cite{DDTIdataset} dataset. Performance is compared against several leading methods in the field, including U-Net \cite{ronneberger2015u}, M-Net \cite{mehta2017m}, Attention Unet \cite{oktay2018attention}, DeeplabV3+ \cite{chen2018encoder}, UNet++ \cite{zhou2018unet++},
BCDU-Net \cite{azad2019bi}, nnUnet \cite{isensee2021nnu},
ARU-GD \cite{Maji2022}, N-Net \cite{nie2022n}, Swin-Unet \cite{cao2023swin} and MShNet \cite{peng2023mshnet}. Table \ref{tab:DDTI} presents the statistical comparison of 
The proposed LSSF-Net has been compared with several advanced techniques. On the DDTI dataset \cite{BUSIdataset}, LSSF-Net achieves a Jaccard index improvement ranging from 5.28\% to 35.95\% over these techniques. Additionally, the performance of LSSF-Net has been tested on thyroid nodule images that present various challenges, including irregular shapes and varying sizes. Figure \ref{fig:Vis_DDTI} presents the visual results of different images of thyroid nodules.

\begin{table}
    \centering
    \caption{Performance comparison of LSSF-Net with various state-of-the-art methods on the thyroid nodule segmentation dataset DDTI.}
    \adjustbox{max width=0.5\textwidth}{%
    \begin{tabular}{lccccc}
    \toprule
    \multirow{2}[4]{*}{\textbf{Method}} & \multicolumn{5}{c}{\textbf{Performance Measures in (\%)}} \\
    \cmidrule{2-6}          & $J_{ind}\uparrow$ & $D_{s}\uparrow$  & $A{cc}\uparrow$ & $S_{n}\uparrow$ & $S_{p}\uparrow$ \\
    \midrule
    U-Net \cite{ronneberger2015u}  & 74.76 & 84.08 & 96.55 & 85.50 & 97.57 \\
    M-Net \cite{mehta2017m} & 79.38 & 86.40 & -     & 75.45 & - \\
    Attention U-Net \cite{oktay2018attention} & 77.37 & 84.91 & -     & 81.70 & - \\
    DeeplabV3+ \cite{chen2018encoder} & 82.66 & 87.72 & -     & 79.54 & - \\
    UNet++ \cite{zhou2018unet++} & 74.76 & 84.08 & 96.55 & 85.50 & 97.57 \\
    BCDU-Net \cite{azad2019bi} & 57.79 & 69.49 & 93.22 & 78.31 & 94.34 \\
    nnUnet \cite{isensee2021nnu} & 80.76 & 88.59 & -     & 85.23 & - \\
    ARU-GD \cite{Maji2022} & 77.07 & 83.64 & 97.94 & 83.80 & 98.78 \\
    N-Net \cite{nie2022n} & 88.46 & 92.67 & -     & 91.94 & - \\
    Swin U-Net \cite{cao2023swin} & 75.44 & 84.86 & 96.93 & 86.42 & 97.98 \\
    MShNet \cite{peng2023mshnet} & 73.43 & 75.01 & -     & 82.21 & - \\
    \midrule
    \textbf{LSSF-Net} & \textbf{93.74} & \textbf{96.72} & \textbf{99.27} & \textbf{96.70} & \textbf{99.52} \\
    \bottomrule
    \end{tabular}%
    }
    \label{tab:DDTI}
\end{table}

For breast cancer segmentation, the performance of LSSF-Net is evaluated on the publicly available BUSI dataset \cite{BUSIdataset}. Performance comparisons are made with multiple state-of-the-art methods, including U-Net \cite{ronneberger2015u}, FPN \cite{lin2017feature}, DeeplabV3+ \cite{chen2018encoder}, ConvEDNet \cite{lei2018segmentation}, UNet++ \cite{zhou2018unet++}, BCDU-Net \cite{azad2019bi}, BGM-Net \cite{wu2021bgm}, ARU-GD \cite{Maji2022}, and Swin-Unet \cite{cao2023swin}.
Table \ref{tab:BUSI} presents the statistical comparison of the proposed LSSF-Net with the state-of-the-art methods. Compared to state-of-the-art methods, the Jaccard index of the proposed LSSF-Net is improved by 15.83\%--25.22\% on the BUSI dataset \cite{BUSIdataset}. The proposed LSSF-Net is also evaluated on breast cancer images with various challenges such as irregular shapes and varying sizes. Figure \ref{fig:Vis_BUSI} presents the visual results of different challenges in breast cancer segmentation.

The proposed LSSF-Net delivered superior segmentation results, closely aligning with the ground truth data, even for thyroid nodule images exhibiting diverse sizes and irregular shapes on the BUSI and DDTI datasets, respectively.

\subsubsection{Computational Complexity Analysis}

\begin{figure}
    \centering
    \includegraphics[width=0.5\textwidth]{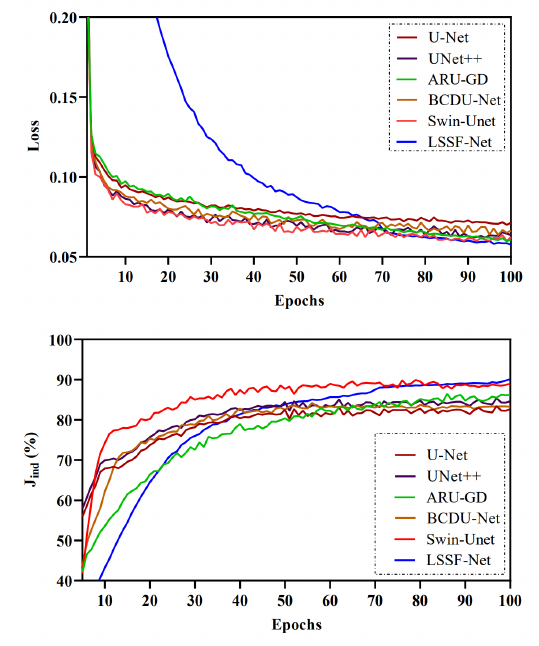}
\caption{Comparison of validation loss and validation Jaccard index during training on ISIC 2017 Dataset}
    \label{fig:validation}
\end{figure}

In this section, we conduct a comprehensive analysis of the computational requirements associated with the LSSF-Net. 
LSSF-Net stands out for its computational efficiency compared to other SOTA models. It converges more quickly in training loss and achieves the highest Jaccard index scores in 100 epochs. Its lightweight architecture requires less GPU memory and supports larger batch sizes, improving scalability and efficiency in medical image analysis.
The graph presented in Figure \ref{fig:validation} provides information on the training loss trajectory of our proposed model compared to alternative algorithms in 100 epochs. Initially, our model exhibited a relatively higher training loss, suggesting a slow start. However, as training progressed, it demonstrated a consistent trend of improvement, steadily reducing loss over successive epochs. This indicates the model's ability to learn from the provided medical dataset and refine its segmentation capabilities over time. At the end of the training period, our model achieved a significantly lower training loss compared to competing algorithms, highlighting its ability to capture and represent the underlying patterns in the data effectively. The computational comparison, presented in Table \ref{tab:Computations}, highlights the efficiency and effectiveness of the LSSF-Net approach.

\begin{table}
  \centering
  \caption{Analysis of Computational Complexity for LSSF-Net, with all evaluations performed on an image resolution of $256 \times 256$.}
 \adjustbox {max width=0.5\textwidth}{%
    \begin{tabular}{lccc}
    \toprule
    \multirow{2}[4]{*}{\textbf{Method}} & \multicolumn{3}{c}{\textbf{Computational Analysis}} \\
\cmidrule{2-4}          & \textbf{Param (M) $\downarrow$} & \textbf{FLOPs (G) $\downarrow$} & \textbf{Inference Time (ms) $\downarrow$} \\
    \midrule
    U-Net \cite{ronneberger2015u} & 32.9  & 33.39 & 28.87 \\
    UNet++ \cite{Zhou2018} & 34.9  & 35.6  & 31.3 \\
    ARU-GD \cite{Maji2022} & 33.3  & 33.93 & 29.49 \\
    DeepLabv3 \cite{chen2017rethinking} & 37.9  & 33.89 & 29.62 \\
    DenseASPP \cite{yang2018denseaspp} & 33.7  & 57.88 & 50.39 \\
    BCDU-Net \cite{azad2019bi} & 28.8  & 38.22 & 28.07 \\
    Swin U-Net \cite{cao2023swin} & 29    & 25.4  & 25.6 \\
    \midrule
    \textbf{LSSF-Net} & \textbf{0.81} & \textbf{3.1} & \textbf{13.7} \\
    \bottomrule
    \end{tabular}%
    }
  \label{tab:Computations}%
\end{table}%

Specifically, the LSSF-Net proposal showcases superior computational efficiency, notably in its significantly reduced number of learnable parameters. LSSF-Net outperforms other algorithms in terms of parameter efficiency, boasting a mere 0.81 million parameters. Crucially, this enhanced efficiency does not compromise the expected top-tier performance in medical imaging analyses. LSSF-Net successfully strikes a balance between computational efficiency and exceptional segmentation results. Furthermore, LSSF-Net requires only 3.1 billion floating point operations, accompanied by a reduced inference time of 13.7 milliseconds. This compactness simplifies the deployment and use of the LSSF-Net method in real clinical settings. Due to its smaller size, the model is more efficient and effective in medical imaging analyses, making it easier to integrate and use in real-time implementations.

While our analysis underscores LSSF-Net's computational efficiency, there are several promising avenues for further enhancement and deployment in real-time or resource-constrained environments. Implementing quantization techniques such as float16, int16, and int8 can significantly reduce model weights and computational requirements, making LSSF-Net more suitable for deployment on devices with limited resources, including CAD systems and mobile devices. These techniques not only help in minimizing memory usage but also improve inference speed. Additionally, fine-tuning LSSF-Net on different modalities could enhance its versatility, enabling it to adapt to various industrial and enterprise-level applications. For cloud-based solutions, these optimizations allow LSSF-Net to operate effectively with fewer compute units, reducing operational costs while maintaining high performance. This approach facilitates the model's integration into scalable cloud environments and supports a range of applications from real-time medical imaging to large-scale data processing, as well as deployment in mobile and CAD systems where resource constraints are critical.

In contrast, Figure \ref{fig:validation} displays the Jaccard index performance of our model relative to other algorithms on the validation dataset throughout the training process. Despite the initial slower performance, our model showed a remarkable trend of continuous enhancement in Jaccard index scores over the epochs. This consistent improvement was accompanied by a corresponding decrease in validation loss, reflecting the model's growing accuracy and proficiency in segmenting medical images. At the end of the training, our model surpassed the performance of other algorithms, exhibiting superior segmentation results and affirming its efficacy in facilitating precise medical image analysis and diagnosis.

\subsubsection{Potential limitations of LSSF-Net}
LSSF-Net, being a lightweight model optimised for binary class segmentation tasks such as skin lesions, BUSI and DDTI segmentation, is highly efficient and effective in these specific scenarios. However, this efficiency comes at a cost: the model's simplicity and reduced depth make it less suitable for more complex problems involving multiple modalities and classes. In such cases, deeper models like Vision Transformers (ViT), which are inherently designed to handle complex and multiclass classification tasks, tend to perform better. Therefore, while LSSF-Net excels in targeted applications, its lightweight architecture may not be sufficient to manage the complexities of multimodalities and multiclass scenarios where greater model depth and sophistication are required.
\subsubsection{Future Work }
Future research could focus on extending LSSF-Net to support multiclass segmentation and multimodalities such as fusion models. This involves developing a single model capable of handling multiple modalities, which would enhance its applicability to various medical imaging and industrial scenarios. By integrating information from different sources, such as combining MRI and CT scans in medical imaging, the model could provide more comprehensive and accurate analyses. This direction not only broadens the scope of LSSF-Net but also addresses the growing need for versatile models in complex real-world applications.

\section{Conclusions}
In conclusion, our research presents a significant advancement in the field of skin lesion segmentation, showcasing the effectiveness of the proposed LSSF-Net architecture. Through extensive experimentation and evaluation, we have demonstrated the robustness and generalisability of LSSF-Net in accurately delineating skin lesions from medical images. The results obtained on benchmark datasets affirm the superior performance of LSSF-Net compared to existing segmentation methods, both in terms of accuracy and computational efficiency. Incorporation of convolutional and recurrent neural network modules has been proven to be instrumental in capturing intricate spatial dependencies and contextual information, leading to improved segmentation outcomes.

Furthermore, the versatility of LSSF-Net is evident in its consistent performance across various skin types and lesion characteristics, highlighting its potential for real-world applications in computer-aided diagnosis of dermatological conditions. The presented findings contribute to ongoing efforts to improve the precision and speed of diagnostic tools in dermatology. As we look ahead, there remains room for future exploration and refinement of the LSSF-Net architecture. The integration of additional data sources and the exploration of transfer learning techniques could further amplify the network's capabilities. Additionally, collaboration with healthcare professionals for real-world validation will be crucial to establishing the practical utility of LSSF-Net in clinical settings.

In summary, the strides made in this research underscore the promising prospects of LSSF-Net in advancing the state of the art in skin lesion segmentation, with implications for improved diagnostic accuracy and patient care in dermatology.

\end{document}